\documentclass[10pt,twocolumn,letterpaper]{article}

\usepackage[pagenumbers]{cvpr} 

\usepackage[dvipsnames]{xcolor}

\usepackage{cite}
\usepackage[load-configurations=version-1]{siunitx}
\usepackage{subcaption} 

\usepackage{amsmath,amsfonts,bm}

\def\eqref#1{equation~\ref{#1}}

\def\1{\bm{1}}

\newcommand{\hx}{\hat{x}}
\newcommand{\ho}{\hat{o}}
\newcommand{\hO}{\hat{O}}
\newcommand{\tx}{\tilde{x}}
\newcommand{\tio}{\tilde{o}}
\newcommand{\ta}{\tilde{a}}

\DeclareMathAlphabet{\mathsfit}{\encodingdefault}{\sfdefault}{m}{sl}
\SetMathAlphabet{\mathsfit}{bold}{\encodingdefault}{\sfdefault}{bx}{n}

\def\gF{{\mathcal{F}}}
\def\gG{{\mathcal{G}}}

\def\gL{{\mathcal{L}}}

\def\sF{{\mathbb{F}}}

\def\sR{{\mathbb{R}}}

\newcommand{\pdata}{p_{\rm{data}}}

\newcommand{\E}{\mathbb{E}}

\newcommand{\normlone}{L^1}

\newcommand\norm[1]{\left\lVert#1\right\rVert}
\newcommand{\image}{\text{image}}
\newcommand{\enc}{\text{enc}}
\newcommand{\dec}{\text{dec}}

\usepackage[pagebackref,breaklinks,colorlinks]{hyperref}
\usepackage{subcaption} 
\usepackage{arydshln}
\newcommand{\imgcell}[1]{\includegraphics[scale=0.5]{#1}}

\usepackage[capitalize]{cleveref}
\crefname{section}{Sec.}{Secs.}
\Crefname{section}{Section}{Sections}
\Crefname{table}{Table}{Tables}
\crefname{table}{Tab.}{Tabs.}

\definecolor{cvprblue}{rgb}{0.21,0.49,0.74}

\def\paperID{89} 
\def\confName{CVPR}
\def\confYear{2024}

 \title{Action-conditioned video data improves predictability}
\author{Meenakshi Sarkar\\
Indian Institute of Science\\
Bengaluru, India\\
{\tt\small meenakshisar@iisc.ac.in}
\and
Debasish Ghose\\
Indian Institute of Science\\
Bengaluru, India\\
{\tt\small dghose@iisc.ac.in}
}
\begin{document}
\maketitle
\begin{abstract}
Long-term video generation and prediction remain challenging tasks in computer vision, particularly in partially observable scenarios where cameras are mounted on moving platforms. The interaction between observed image frames and the motion of the recording agent introduces additional complexities. To address these issues, we introduce the Action-Conditioned Video Generation (ACVG) framework, a novel approach that investigates the relationship between actions and generated image frames through a deep dual Generator-Actor architecture. ACVG generates video sequences conditioned on the actions of robots, enabling exploration and analysis of how vision and action mutually influence one another in dynamic environments. We evaluate the framework's effectiveness on an indoor robot motion dataset which consists of sequences of image frames along with the sequences of actions taken by the robotic agent, conducting a comprehensive empirical study comparing ACVG to other state-of-the-art frameworks along with a detailed ablation study.
\end{abstract}    
\section{Introduction}
\label{sec:intro}
The significance of prediction in the context of intelligence is widely recognized \cite{bubic}. While self-attention-based transformer models have revolutionized language prediction and generation \cite{Vaswani_attention2017}, the same level of advancement has not been achieved in the domain of video and image prediction models. However, long-term reliable video prediction is a valuable tool for decoding essential information about the environment, providing a data-rich format that can be leveraged by other learning frameworks like policy gradient \cite{Kaiser2020} or planning algorithms  \cite{Hafner2019}. However, the complex interactions among objects in a scene pose significant challenges for long-term video prediction \cite{finn, finn2, mathieu, villegas, GaoCVPR2020, villegasNeurIPS2019}.
Like every dynamical system, the spatio-temporal dynamics embedded in video data can be represented  as:
\begin{equation}
    \dot{x}_t=f(x_t,a_t)
    \label{eq:vid_dynamics}
\end{equation}
where $x_t$ is the image or pixel state at time step $t$, $a_t$ is the control action or movement of the recording camera,  and $\dot{x}_t$ represents the velocity of the image/pixel state $x_t$. When the camera is stationary ($a_t=0$), \cref{eq:vid_dynamics} represents the classical video prediction and generation problem which has been the primary focus of various previous works such as \cite{srivastava, oh, vondrick, finn, mathieu, villegas, xu, WichersICML2018} over the past decade. These frameworks often utilize optical flow and content decomposition approaches to generate pixel-level predictions. Adversarial training is commonly employed to enhance the realism of the generated images. However, with the increasing availability of computational power, there is a recent trend towards generating high-fidelity video predictions using various generative architectures, such as Generative Adversarial Networks (GAN) and Variational Autoencoders (VAE) \cite{LiangICCV2017, Denton, BabaeizadehICLR2018, LeeICLR2018,CastrejonICCV2019, GaoCVPR2020,villegasNeurIPS2019}. 

However, in recent years, there has been considerable interest in the vision community in predicting video with a dynamic background \cite{GaoCVPR2020,villegas,villegasNeurIPS2019,sarkar2021} where the camera is also moving ($a_t\neq 0$ in \cref{eq:vid_dynamics}). These problems are frequently encountered in autonomous robots and cars and are referred to as prediction in partial observability. Prediction in partially observable scenarios are particularly challenging due to the intrinsic relationship between the movements of the camera and the observed image frame. The dynamics of the recording platform plays a crucial role in conditioning each pixel of data observed by the camera sensor \cite{villegasNeurIPS2019}, \cite{sarkar2021, acpnet2023}. 
%
%
While \cref{eq:vid_dynamics} expresses the relationship between the velocity of pixel data, image frame $x_t$ and control action of the recording camera in a very simplified format, the fundamental philosophy that models such systems remains the same. For example,
 \cite{sarkar2021} proposed a Velocity Acceleration Network (VANet) that modeled the interaction between the camera action ($a_t$) and the observed image frame ($x_t$) with second-order optical flow maps. Villegas et al. \cite{villegasNeurIPS2019} attributed this interaction to stochastic noise and showed that, with large enough parametric models with higher computing capabilities,  longer accurate predictions can be made with video data.  With the success of transformer models in natural language problems,  there has been current interest in designing efficient visual transformers \cite{ViT2021} for video prediction tasks \cite{video_transformer2022, Gao_2022_CVPR_ViT}. An interesting work is by Gao et al. \cite{Gao_2019_ICCV} which addressed the problem arising from partial observability in KITTI dataset \cite{kitti} by disentangling motion-specific flow propagation from motion-agnostic generation of pixel data for higher fidelity. 

However, none of these works explicitly model or incorporate the action data $a_t$ of the recording platform. This can be partly attributed to the fact that none of the existing datasets  \cite{kitti},  \cite{KITTI-360},  \cite{A2D2} provide any synchronized action data along with the image frames. A recent study  \cite{acpnet2023} introduced the RoAM or Robot Autonomous Motion dataset, which includes synchronized robot action-image data. This dataset provides stereo image frames along with time-synchronized action data from the data-recording autonomous robot. They showed that their Action Conditioned Prediction Network (ACPNet) incorporating the past action of the video data recording robot explicitly into the prediction model leads to better performance and accuracy.

 While ACPNet introduced the idea of explicitly conditioning the generated future image frames on the past action sequences of the recording platform, their model assumed constant control action for the entire prediction horizon $T$. While this simplification may be acceptable for short prediction horizons, for longer durations (larger $T$), this assumption can result in decreased accuracy in predictions. While predicted image frames rely on the actions taken by the mobile agent, actions by the agent often depend on the current system state, especially in partially observable scenarios arising in autonomous cars and indoor mobile robots like in the case of RoAM dataset. The actions taken by robots in these scenarios are often continuous, and smooth and have a direct relationship to the current state of the system and can be modelled given sufficient data. 
 
 We propose that accurate approximations of future control actions can enhance predictions of the future image state and vice versa. In this paper, we introduce an Action-Conditioned Video Generation (ACVG) framework, that models this dual dependency between the state and action pair of partially observable video data through our Generator-Actor Architecture. In addition to a generator architecture that considers the current and past platform actions, ACVG features an actor-network that predicts the next action, $\tilde{a}_{t+1}$, based on the current observed/predicted image state to highlight the influence between modeling the system controller and the system dynamics derived from video frames. 
 Our dual network implicitly encodes the dynamics of the recording platform and its effect on predicted image frames by explicitly modeling and predicting the upcoming course of action through the actor-network.

The rest of the paper is organized as follows: The framework of the Generator and Actor network of the proposed ACVG framework is proposed in  Section \ref{sec:acvg} along with loss functions and the training process. In Section \ref{sec:dataset}, we  explain why the RoAM dataset is an ideal choice to validate our proposed hypothesis.  The experimental setup for training and testing ACVG is given in Section \ref{sec:dataset}. The performance analysis and ablation study of ACVG, compared to other frameworks: ACVG with fixed action (ACVG-fa), ACPNet and VANet  are given in Section \ref{sec:results}. Finally, Section \ref{sec:conclusion} concludes the paper with a few remarks regarding the possible future research directions.

\section{Action Conditioned Video Generation}\label{sec:acvg}

\begin{figure}[t]
\centering
  \includegraphics[width=0.9\linewidth]{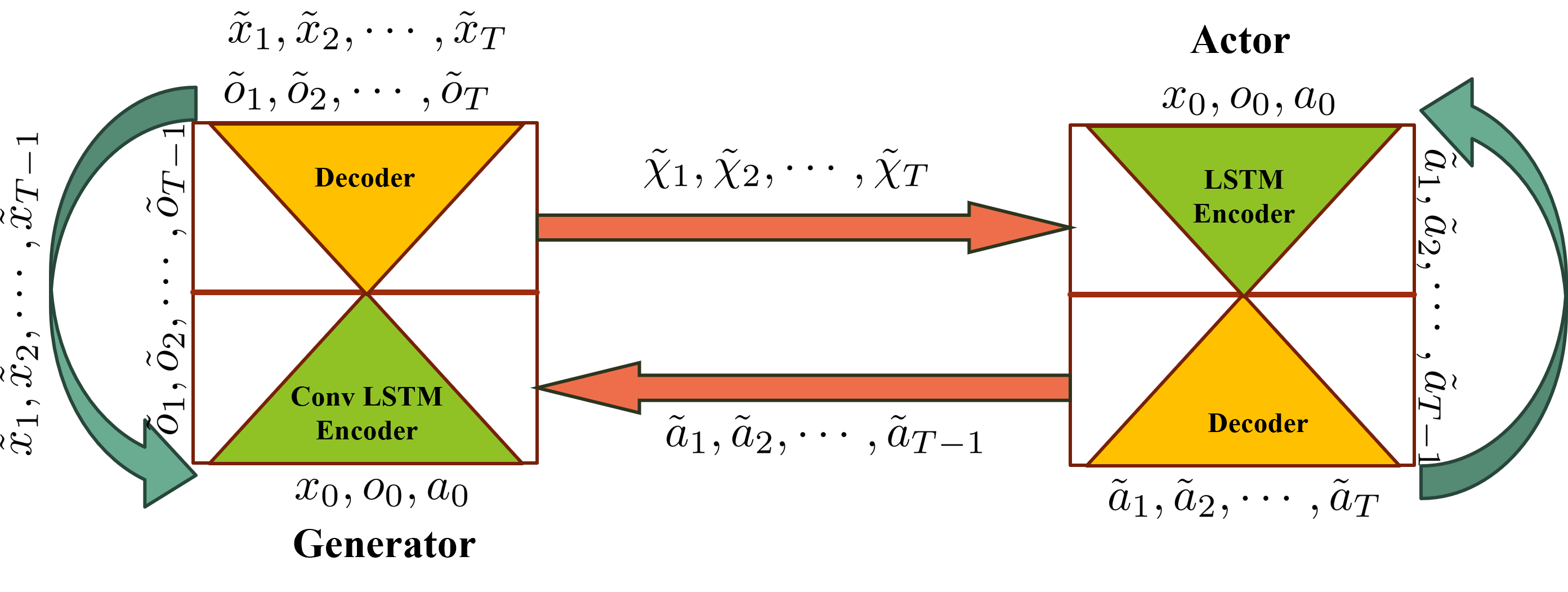}
  \caption{The interdependent training loop of the Generator and Actor network of ACVG.  }
\label{fig:algo_interplay}
\end{figure}

Let the current image frame be denoted by  $x_t \in \sR^{[\image_h\times\image_w\times c]}$ where $\image_h$, $\image_w$, and $c$ denote the height, width, and channel length of the image, respectively. In the case of color images like ours $c=3$. We define the first-order pixel flow map  $o_t \in \sR^{[\image_h\times\image_w\times c]}$ at time step $t$ as the following: 
\begin{equation}
    o_t=x_t-x_{t-1}
    \label{eq:flow_map}
\end{equation}
The normalized action at time-step $t$ is denoted as  $a_t \in \sR^m$ where $m$ is the dimension of the action space of the mobile agent on which the camera is mounted. This normalization approach is useful for generalizations needed due to the potential variations in actuator limits among different mobile platforms, ensuring that the inference framework can scale effectively, accommodating diverse actuator ranges.

Our objective here is the predict to future image and action state pair $(x_t,a_t)$ from $t=1$ to $t=T$, where $T$ is the prediction horizon, given the initial image, flow, and action state as $x_0$, $o_0$, and $a_0$, respectively, at the beginning of the prediction process at time-step $t=0$.  For a deterministic dynamical system this would mean an approximation of the following equations:
\begin{align}
    x_{t+1}=x_t+\gF(x_t,o_t,a_t),&\hspace{1em} \text{given }(x_0,o_0,a_0) \label{eq:deterministic_xdot1}\\
    a_{t+1}= \gG(x_{t+1},o_{t+1})&
    \label{eq:deterministic_xdot2}
\end{align}
where $\gF$ and $\gG$ denote the dynamics the system and the controller, respectively, which are both unknown. In the case of video prediction, the approximation of \cref{eq:deterministic_xdot1,eq:deterministic_xdot2} becomes involved and complex due to the high dimensionality of the data and the nature of the complex interactions between the motion of the camera and the motion of the objects captured on the image plane of the camera.
We map the high-dimensional image information in $x_t$ and $o_t$ on a low-dimensional feature space as $\hx_t$ and $\ho_t$ using encoder architectures. In the low dimensional space, we can expect to observe interaction behaviour between $\hx_t$, $\ho_t$ and $a_t$ similar to \cref{eq:deterministic_xdot1,eq:deterministic_xdot2}, although these interactions might not be in an affine form as suggested in \cref{eq:deterministic_xdot1}. 

We approximate these dynamic interactions and predict future image frames with the   
proposed Action Conditioned Video Generation (ACVG) framework. ACVG is a deep dual network consisting of a Generator and an actor network. The objective of the generator network is to loosely approximate \cref{eq:deterministic_xdot1} whereas the actor network tries to approximate $\gG$ from \cref{eq:deterministic_xdot2}. 
\cref{eq:deterministic_xdot1,eq:deterministic_xdot2} exhibit a mutual dependence. Similarly, within the Action-Conditioned Video Generation (ACVG) framework, the generator and actor networks share an interconnected relationship, as shown in Figure \cref{fig:algo_interplay}, where  at each time step $t$, the generator network uses the predicted states $\tx_t$ and $\tio_t$, along with the anticipated action state $\ta_t$ generated by the actor network, to forecast the subsequent image frame $\tx_{t+1}$. This forecasted $\tx_{t+1}$ is subsequently used by the actor network to project the likely action at time $t+1$, denoted as $\ta_{t+1}$, which is then fed back into the generator network. This iterative cycle continues till the prediction horizon  $T$. In this paper $(\tilde{.})$ denotes predicted states.

The initial states, denoted by $x_0$ and $o_0$, carry a broader interpretation beyond representing the image and flow map  at time step $t=0$. We employ an augmented notion of the initial state, where $x_0$ signifies a collection of past observations spanning from $t=-n+1$ to $t=0$ as $x_0=\{x_{t=0},x_{t=-1},\cdots,x_{t=(-n+1)}\}$. Likewise, for flow maps, we have $o_0=\{o_{t=0},o_{t=-1},\cdots,o_{t=(-n+1)}\}$. It is worth noting that the computation of flow maps, in accordance with  \cref{eq:flow_map}, considers ${o_{(-n+1)}=0}$, as it assumes that, at the commencement of the observation at $t=-n+1$, the environment was in a static state. This convention extends to the initial action state as well, and $a_0=\{a_{t=0},a_{t=-1},\cdots,a_{t=(-n+1)}\}$. In this specific scenario, the minimum number of past observations required is $n_{\min}=2$, as having at least one observed flow map at $t=0$ is essential. However, from the experimental findings, it appears  more advantageous to maintain $n\geq 3$ for more robust and improved inference within the generative framework.
\begin{figure*}[ht]
\centering
\subcaptionbox{\label{fig:acvg_generator}}{\includegraphics[height=6.2cm]{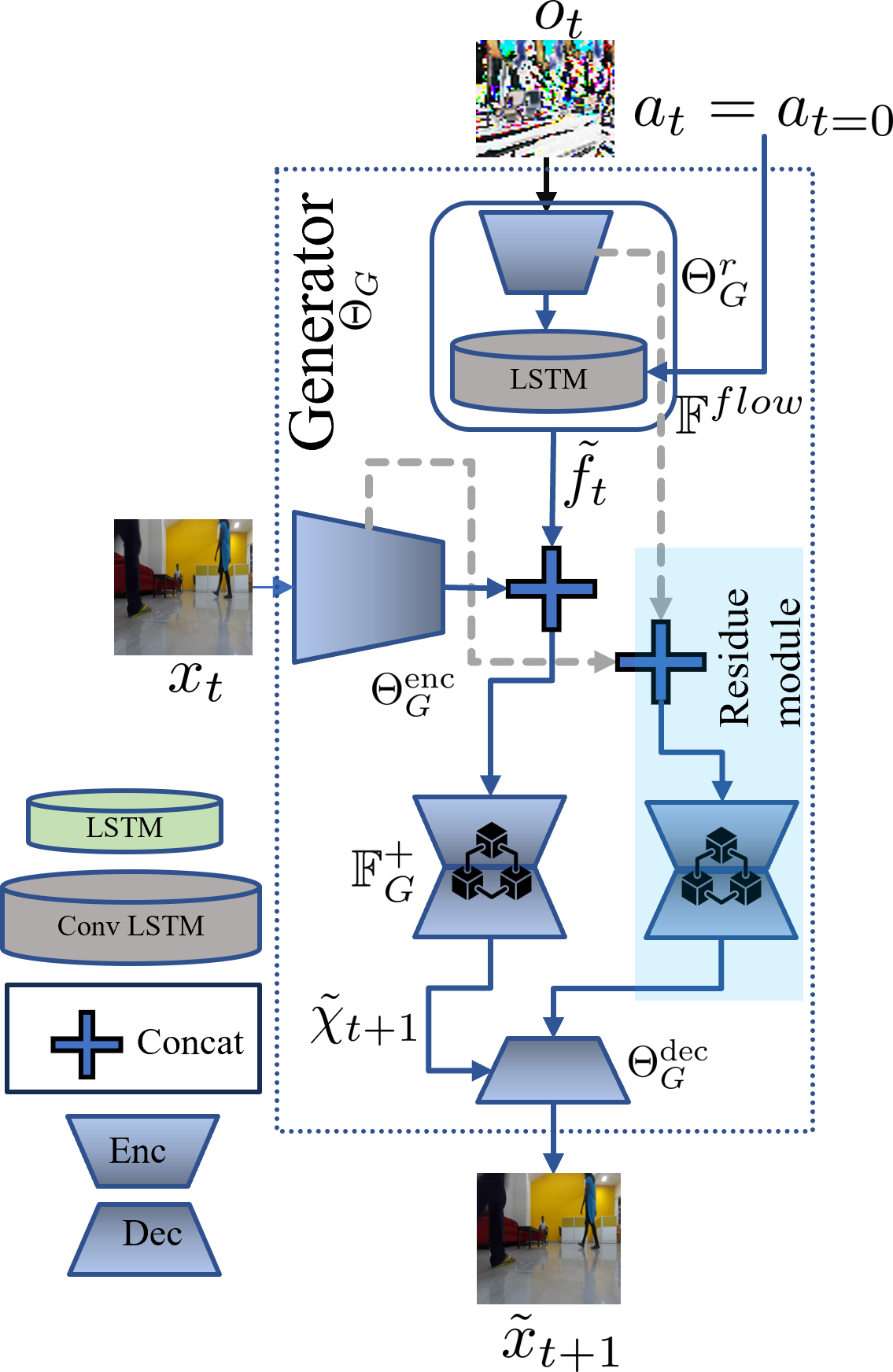}}%
\hfill 
\subcaptionbox{\label{fig:acvg_actor}}{\includegraphics[height=6.2cm]{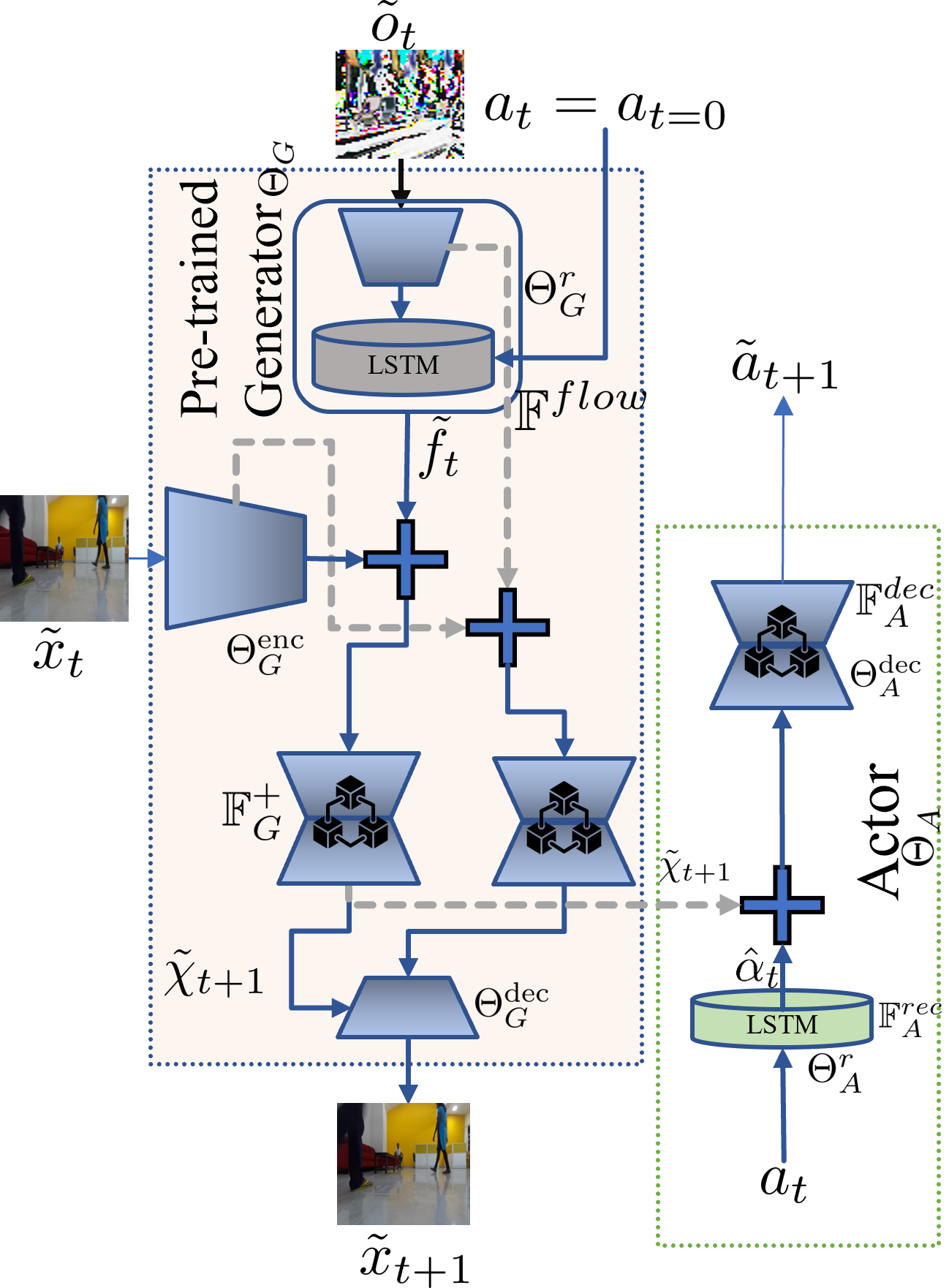}}%
\hfill 
\subcaptionbox{\label{fig:acvg_dual}}{\includegraphics[height=6.2cm]{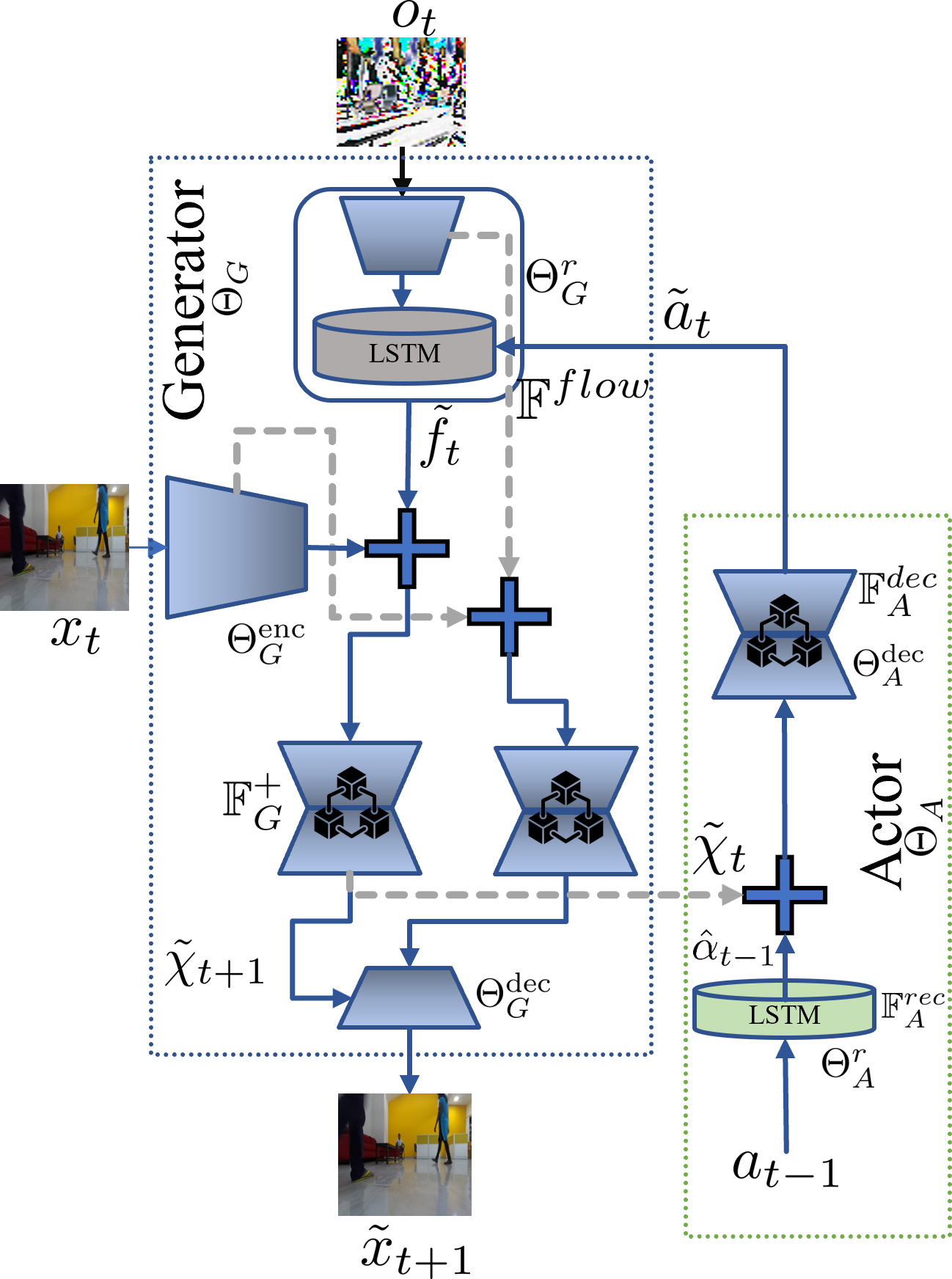}}
\caption{Architecture of ACVG consists of dual networks: Generator and Actor. Fig. \ref{fig:acvg_generator} shows the architecture of the generator network alone during the generator training phase. During this phase, $\beta=1$ and $\gamma=0$ for the training loss in \cref{eq:loss_ln_expand} and $a_t$ is constant. Fig. \ref{fig:acvg_actor} shows the dual configuration of the Generator-Actor framework during the actor training phase. During this phase the weights of the pre-trained generator network is kept frozen. The generator network is used in the inference mode with $\beta=0$ and $\gamma=1$ in \cref{eq:loss_ln_expand}. Finally \cref{fig:acvg_dual} shows the configuration of the network in dual training mode when $\beta=1$ and $\gamma=1$.It's important to emphasize that in the dual training phase, as depicted in Figure \cref{fig:acvg_dual}, we employ a delayed actor-network to provide the approximate current action $\ta_t$ to the generator network. This delay is necessary because we operate with a causal model, wherein the actor network generates the predicted action $\ta_{t+1}$ based on the observation of state $\tilde{\chi}_{t+1}$.}
    \label{fig:ACVG}
\end{figure*}
\subsection{Generator Network}\label{generator}
We use stacked recurrent units like Convolutional LSTMs \cite{convlstm} and LSTM cells \cite{lstm} in case of the generator and actor network, respectively, to iteratively approximate the dynamics from the image and action data. The detailed dual generator-actor network architecture of ACVG is shown in \cref{fig:ACVG}. The Generator network, is parameterized by $\Theta_G$. The parametric space of $\Theta_G$ is further divided into 3 groups: $\Theta_G=[\Theta^{\enc}_G, \Theta^r_G,\Theta^{\dec}_G ]$ where $\Theta^{\enc}_G$ denotes the encoder network that maps the image $x_t$ and flow maps $o_t$ at time $t$ into the low dimensional manifold. This is expressed as:
\begin{equation}    x_t\xrightarrow[\Theta^{\enc}_G]{\mbox{Encoder}} \hx_t, \hspace{2em }  o_t\xrightarrow[\Theta^{\enc}_G]{\mbox{Encoder}} \ho_t
    \label{eq:hx_t}
\end{equation}
where $\Theta^r_G$ represent the recurrent part of the network that is designed with stacked convolutional LSTM layers to approximate the spatio-temporal dynamics from the recursive first order flow maps $\ho_t$ similar to \cite{villegas},\cite{sarkar2021}. However, unlike these works, we use an augmented flow map $\hO_t$, where we concatenate the flow map $\ho_t$ with the normalized action $a_t$ at time $t$ as follows:
\begin{equation}
    \hat{O}_t=[\ho_t, a_t]
    \label{eq:O_hat}
\end{equation}
This augmented flow map, $\hO_t$ is then fed to the flow or the recursive LSTM module of the generator to approximate the motion kernel $\tilde{f}_t$ as shown in \cref{fig:acvg_generator,fig:acvg_dual}. The flow module $\sF^{flow}$ approximates the temporal dynamics from the observed image flow data and the control actions taken by the recording agent $a_t$ and is similar to the system dynamics function $\gF$ in \cref{eq:deterministic_xdot1}. However, $\sF^{flow}$ approximates the dynamics in the parametric space of $\Theta^r_G$, given as:    
\begin{equation}
    \tilde{f}_t=\sF^{flow}_{G}(\hO_t|\hO_{1:t-1},\Theta^r_G, x_0,o_0,a_0)
    \label{eq:MoEnc}
\end{equation} 
Once  the motion kernel $\tilde{f}_t$ in \cref{eq:O_hat,eq:MoEnc} is evaluated, it is  combined with the current image state in the feature space of $\hx_t$ to generate $\tilde{\chi}_{t+1}$:
\begin{equation}    \tilde{\chi}_{t+1}=\sF^{+}_{G}(\hx_t,\tilde{f}_t)
    \label{eq:chi_t+1}
\end{equation}
The generated $\tilde{\chi}_{t+1}$ is then decoded (\cref{fig:acvg_generator}) and mapped back into the original image dimension to generate the predicted image frame $\tx_{t+1}$ as:
\begin{equation}
    \tilde{x}_{t+1}\xleftarrow[\Theta^{\dec}_G]{Dec} \tilde{\chi}_{t+1}
    \label{eq:x_t+1}
\end{equation}
where $\Theta^{\dec}_G$ denotes the parametric space of the decoder network. Clubbing the feed-forward components $\Theta^{\enc}_G$ and $\Theta^{\dec}_G$ together  as $\sF^{\overline{flow}}_G$,  \cref{eq:hx_t,eq:O_hat,eq:MoEnc,eq:chi_t+1,eq:x_t+1} can be written as :
\begin{equation}
\tilde{x}_{t+1}=\sF^{flow+\overline{flow}}_G(\hat{x}_t, \hat{O}_t|\Theta_G,\hO_{1:t-1},x_0,o_0,a_0 )
    \label{eq:future_x1}
\end{equation}
which can be further re-arranged as:
\begin{equation}
    \tilde{x}_{t+1}=\sF_{G}(x_t, o_t, a_t|\Theta_G,x_{0:t-1},o_{0:t-1},a_{0:t-1})
    \label{eq:predict_x}
\end{equation}
However, during inference, the generator network iteratively uses previously predicted image $(\tx_t)$, flow $(\tio_t)$ and action $(\ta_t)$ states, and hence \cref{eq:predict_x} can be rewritten as:
\begin{equation}
    \tilde{x}_{t+1}=\sF_{G}(\tx_t, {\tio}_t, \ta_t|\Theta_G,\tx_{1:t-1},\tio_{1:t-1},\ta_{1:t-1}, X_0)
    \label{eq:predict_x_infr}
\end{equation}
where $X_0=[x_0,o_0,a_0]$. From the expression for $\tilde{x}_{t+1}$ on the right-hand side (RHS) of \cref{eq:predict_x_infr}, it is clear that the predicted image frames are conditioned on the predicted control actions $\ta_t$ approximated by the actor network. We also use residual or skip connections similar to \cite{villegas,sarkar2021}, shown in \cref{fig:acvg_generator} to improve accuracy.
While Sarkar et. al \cite{acpnet2023} used similar augmentation of action with the image $x_t$ for their video generation network, the proposed ACVG is the first framework that explores the augmented flow map $\hO_t$ as a means to not only improve the image prediction accuracy but also to capture and predict probable future actions executed by the mobile agent, facilitated by the dual actor network which we discuss next.

\subsection{Actor Network}\label{subsec:actor}
 The actor network is parameterized by $\Theta_A$ and shown in \cref{fig:acvg_actor,fig:acvg_dual}. Similar to the generator network, it comprises both a recurrent module and a decoder module. The parametric space $\Theta_A$ is further divided into two components: $\Theta_A=[\Theta^r_A, \Theta^{\dec}_A]$. Here, $\Theta^r_A$ pertains to the recurrent module of the actor network, while $\Theta^{\dec}_A$ are the parameters of the decoder module responsible for generating the predicted action $\ta_{t+1}$. This predicted action is contingent on the knowledge of the latent image state $\hx_{t+1}$ and the flow map $\ho_{t+1}$. This formulation is parallel to the expression for the control action $a_{t+1}$ in  \cref{eq:deterministic_xdot2} for dynamical systems. 

 The recurrent module of the network is built with LSTMcell \cite{lstm} architectures that compute the temporal dynamics encoded in the past action data and can be represented as: \begin{equation}    \hat{\alpha}_t=\sF^{rec}_A(a_t|\Theta^r_A, a_{0:t-1})
    \label{eq:actor_rec}
\end{equation}
The recurrent module $\sF^{rec}_A$ in \cref{eq:actor_rec}  assumes the action $a_t$ taken by the robot to be a smooth and continuous function of the image $x_t$ and flow state $o_t$ and the robot does not make any sudden or jump movements. Thus, we use a recurrent architecture to capture these smooth temporal dynamics (\cref{fig:acvg_actor}).
With $\hat{\alpha}_t$, the predicted action  at time step $t+1$ can be generated with latent space information of the future image state $\hx_{t+1}$ and flow map $\ho_{t+1}$ as:
\begin{equation}    \tilde{a}_{t+1}=\sF^{dec}_A(\hx_{t+1},\ho_{t+1}, \hat{\alpha}_t|\Theta^{\dec}_A)
    \label{eq:action_pred_T}
\end{equation}
However, in reality, during inference, we do not have access to the true state information on $\hx_{t+1}$ and $\ho_{t+1}$. Instead, we can have an approximate estimate of the future image flow map $\tilde{\chi}_{t+1}$ generated by the generator network as expressed in \cref{eq:chi_t+1}. Thus we can express \cref{eq:action_pred_T} as the following:
\begin{equation}   \tilde{a}_{t+1}=\sF^{dec}_A(\tilde{\chi}_{t+1},\hat{\alpha}_t|\Theta^{\dec}_A)
    \label{eq:action_pred}
\end{equation}
  \cref{eq:action_pred} shows how the predicted action state $\tilde{a}_{t+1}$ relies on the latent image flow maps $\tilde{\chi}_{t+1}$ approximated by the Generator (\cref{fig:acvg_actor}). This illustrates the
  relationship between the Generator and Actor networks, as shown in  \cref{fig:algo_interplay} and \cref{fig:ACVG}. Without  loss of generality, we can write \cref{eq:action_pred} as :
\begin{equation}    \tilde{a}_{t+1}=\sF^{rec+dec}_A(a_t, \tilde{\chi}_{t+1}|\Theta^f_A,\Theta^r_A, a_{0:t-1})
    \label{eq:actor_comb1} 
\end{equation}
During inference, when both the image and action states are being iteratively approximated, \cref{eq:actor_comb1} becomes, 
\begin{equation}
    \tilde{a}_{t+1}=\sF_{A}(\tx_{t+1},\tio_{t+1},\ta_t|\Theta_A,\ta_{1:t-1},a_0)
    \label{eq:actor}
\end{equation}
Upon examining  \cref{eq:predict_x_infr} and \cref{eq:actor}, it becomes apparent that because of the mutually dependent relationship between the generator and actor networks, the prediction accuracy of the generator network has a direct impact on the effectiveness of the actor network, and vice versa. If the actor network is poorly designed and trained, it can generate inaccurate future action predictions which, in turn, will affect the prediction accuracy of the generator network. This interplay depicted in \cref{fig:algo_interplay}, can lead to a cascading effect of poor decision-making within the system. This interdependency is manifested in the design of our loss function for the integrated ACVG framework and the training procedure for the dual network, which is explianed next.
\subsection{Loss and Training Loop}\label{subsec:loss}
 Our objective is to find the optimum $\Theta^*=\{\Theta_G^*, \Theta_A^*\}$ such that the joint likelihood function of $x_{1:T}$ and $o_{1:T}$ and $a_{1:T}$ is maximised for a given $X_0=[x_0,o_0,a_0]$. This can be expressed as: 
 \begin{equation}
 \begin{aligned}
     \underset{\Theta}{\max}&\gL_{\Theta}(S_{1:T},a_{1:T}|X_0)=\\
     &\sum_{t=1}^{T}\E_{\pdata}\text{ln}p_{\Theta^*}(S_t,a_t|S_{0:t-1},a_{0:t-1})
 \end{aligned} \label{eq:loss_ln}
 \end{equation}
 where $S_t=[x_t,o_t]$ represents the joint system states $x_t$ and $o_t$ and $\text{ln}p(S_t,a_t)$ on the RHS of \cref{eq:loss_ln} represents the joint logarithmic likelihood probability of $(S_t,a_t)$, which  can be further expanded as:
 \begin{equation}
 \begin{aligned}
     \text{ln}&p_{\Theta}(S_t,a_t|S_{0:t-1},a_{0:t-1})=\beta[\text{ln}p_{\Theta_G}(x_t|S_{0:t-1},a_{0:t-1})\\
     &+\text{ln}p_{\Theta_G}(o_t|x_t,S_{0:t-1},a_{0:t-1})]+\gamma\text{ln}p_{\Theta_A}(a_t|S_{0:t},a_{0:t-1})
 \end{aligned} \label{eq:loss_ln_expand}
 \end{equation}
When $\beta=1$ and $\gamma=1$, we obtain the expression for Bayesian inference from \cref{eq:loss_ln_expand}. However, the values of $\beta$ and $\gamma$ vary depending on the specific phases within the training loop, as will be explained later. The first expression on the right-hand side (RHS) of \cref{eq:loss_ln_expand} can be calculated as a penalty on the expected reconstruction loss for the generated image frames from time $t=1$ till $t=T$ and can be expanded as:
\begin{equation}
\begin{aligned}
 \underset{\Theta_G}{\max}\sum_{t=1}^{T}
   &(\text{ln}p_{\Theta_G}(x_t|S_{0:t-1},a_{0:t-1}))\equiv \underset{\Theta_G}{\min} \sum_{t=1}^{T} \big[\norm {{x}_{t} - {\tilde{x}}_{t} }^{\lambda_1}\\
   &+\sum_{i,j}^{w,h}\big(||{x}_{t,i,j}- {x}_{t,i-1,j} |- |{\tilde{x}}_{t,i,j}- {\tilde{x}}_{t,i-1,j} ||^{\lambda_2}+\\
&\hspace{2em}+||{x}_{t,i,j}- {x}_{t,i,j-1} |- |{\tilde{x}}_{t,i,j}- {\tilde{x}}_{t,i,j-1} ||^{\lambda_2}\big)\big]
    \end{aligned}
    \label{eq:L_rec_x}
\end{equation}
Here both ${\lambda_1}$ and ${\lambda_2}$ can take the values  $1$ or $2$. However, for our experiments, we have chosen both $\lambda_1=\lambda_2=1$. Similar to \cref{eq:L_rec_x}, we can expand the likelihood function for the flow map in \cref{eq:loss_ln_expand} as:
\begin{equation}
\begin{aligned}
\underset{\Theta_G}{\max}\sum_{t=1}^{T}
   &(\text{ln}p_{\Theta_G}(o_t|x_t,S_{0:t-1},a_{0:t-1}))\equiv \underset{\Theta_G}{\min} \sum_{t=1}^{T} \big[\norm {{o}_{t} - {\tilde{o}}_{t} }^{\lambda_1}\\
   &\sum_{i,j}^{w,h}\big(||{o}_{t,i,j}- {o}_{t,i-1,j} |- |{\tilde{o}}_{t,i,j}- {\tilde{o}}_{t,i-1,j} ||^{\lambda_2}+\\
&\hspace{2em}+||{o}_{t,i,j}- {o}_{t,i,j-1} |- |{\tilde{o}}_{t,i,j}- {\tilde{o}}_{t,i,j-1} ||^{\lambda_2}\big)\big]
    \end{aligned}
    \label{eq:L_rec_o}
\end{equation}
In \cref{eq:L_rec_x},  in order to evaluate the reconstruction penalty, we have not only calculated the $\normlone$ loss between the ground truth image $x_t$ and predicted image $\tx_t$, but also the spatial gradient loss \cite{villegas},\cite{sarkar2021},\cite{acpnet2023} between the two. The same is done for flow maps $o_t$ and $\tio_t$ in \cref{eq:L_rec_o}. The reconstruction penalty for the action likelihood function, with $\lambda_a=2$, is calculated as follows:
\begin{equation}
\underset{\Theta_A}{\max}\sum_{t=1}^{T}
\text{ln}p_{\Theta_A}(a_t|S_{0:t},a_{0:t-1})\equiv \underset{\Theta_A}{\min} \sum_{t=1}^{T} \norm {{a}_{t} - {\tilde{a}}_{t} }^{\lambda_a}
    \label{eq:lfap_L_rec_a}
\end{equation}
Other than the likelihood loss expressed via \cref{eq:loss_ln,eq:loss_ln_expand,eq:L_rec_x,eq:L_rec_o,eq:lfap_L_rec_a}, we also use a scaled generative adversarial loss $\mathcal{L}_{adv}$, similar to \cite{mathieu}. Due to the averaging effects of the deterministic framework, this adversarial loss is added to the generator loss given by \cref{eq:L_rec_x} and \cref{eq:L_rec_o} to reduce the blurring effects in the prediction and generate more realistic images \cite{villegas},\cite{sarkar2021}. Detailed expression for the adversarial loss along with a short description of the discriminator network is given in the supplementary material.\\
\textbf{Training Loop:} The training process for the ACVG framework comprises three distinct phases: (i) Generator Training, (ii) Actor Training, and (iii) Dual Training. We initially focus on training the Generator network alone for $N_g$ number of iterations as shown in \cref{fig:acvg_generator}. In this phase, we set $\beta=1$ and $\gamma=0$ in \cref{eq:loss_ln_expand}. Once the generator network  reaches an adequate level of proficiency in grasping the spatio-temporal dynamics inherent in the flow and image data, we freeze the network's weights. During the generator training phase, we assume that all future control actions, denoted as $a_{1:T}$, remain constant and equal to the last observed control action at $a_{t=0}$, that is, $a_{1:T}=a_{t=0}$.

After freezing the generator network's weights, the actor network is then trained for a specified number of iterations, $N_a$. During this phase, as shown in \cref{fig:acvg_actor}, the frozen Generator network operates in inference mode, where predictions from the actor network $\ta_{t+1}$ are conditioned on the predicted image flow state $\tilde{\chi}_{t+1}$ from the generator network, as outlined in \cref{eq:action_pred}. Throughout the actor network training phase, we set $\beta=0$ and $\gamma=1$. Subsequently, once the actor network  reaches an adequate level of training, we transition to dual training mode, where both the generator and actor networks are trained together. In this dual training mode, we set $\beta=1$ and $\gamma=1$. This dual training mode spans $N_{\text{dual}}$ iterations and is shown in \cref{fig:acvg_dual}, which shows that in the dual training phase, a delayed actor-network is employed to provide the current action $\ta_t$ to the generator network. This design choice aligns with our utilization of a causal model, where the actor network generates the  action $\ta_{t}$ grounded on the current observed state $\tilde{\chi}_{t}$ (shown with the dotted grey line in \cref{fig:acvg_dual}).
While tuning the hyperparameters $N_g$, $N_a$, and $N_{\text{dual}}$ is necessary, we found it manageable by monitoring individual reconstruction losses from the generator and actor networks. 
\section{RoAM dataset and Experimental Setup }\label{sec:dataset}
We use Robot Autonomous Motion or RoAM dataset \cite{acpnet2023}, which contains synchronized control action data along with the image frames,  to test and benchmark  ACVG against other state-of-the-art frameworks: ACPNet, VANet and MCNet. There are multiple partially observable datasets available in the literature \cite{acpnet2023}. However, RoAM is the only dataset that provides the synchronized control action data along with the image frames to properly test our hypothesis. RoAM is an indoor human motion dataset, collected using a custom-built Turtlebot3 Burger robot. It consists of 25 long synchronized videos and control-action sequences recorded capturing corridors, lobby spaces, staircases, and laboratories featuring frequent human movement like walking, sitting down, getting up, standing up, etc. The dimension of the action data is $m=2$ featuring forward velocity along the body $x$-axis and turn rate about the body $z$-axis. 


The dataset is split into training and test sets with a ratio of 20:5, respectively. Following the training process described in Section \ref{subsec:loss}, ACVG is trained on the RoAM dataset using the ADAM optimizer \cite{adam} with an initial learning rate of 0.0001 and a batch size of 8. 
During the first phase of the training, the Generator network was trained on the entire training dataset for 1000 epochs for generating the pre-training weights as depicted in \cref{fig:acvg_generator}. Once the pre-training is done, we freeze the weights of the generator and train the actor network with generator in the inference mode (\cref{fig:acvg_actor}). We pre-train the actor network on the full training dataset for 500 epochs. Once the pre-training weights for both the network are obtained, both the networks are trained in the dual mode (\cref{fig:acvg_dual}) for the next 1000 epochs. During both training and testing, each of the 25 video sequences was divided into smaller clips of 50 frames. A gap of 10 frames is maintained between each clip to ensure independence between two consecutive sequences.  

The network was trained to predict 10 future frames based on the past 5 image frames of size $128\times128\times3$ and the corresponding history of control actions. During inference, the ACVG framework generates 20 future frames while conditioning on the last 5 known image frames and their corresponding action sequences. We have also trained ACPNet, VANet and MCNet using  similar hyperparameters as described in \cite{sarkar2021}, \cite{acpnet2023}, \cite{villegas}.
Each of these networks was trained for 150,000 iterations on a GTX 3090 GPU-enabled server as mentioned in \cite{sarkar2021} and \cite{acpnet2023}.

\section{Results and Discussion}\label{sec:results}
\begin{figure*}
\centering
\subcaptionbox{\label{fig:roam_quant_vgg16}}{\includegraphics[width=0.25\textwidth,height=3.6cm]{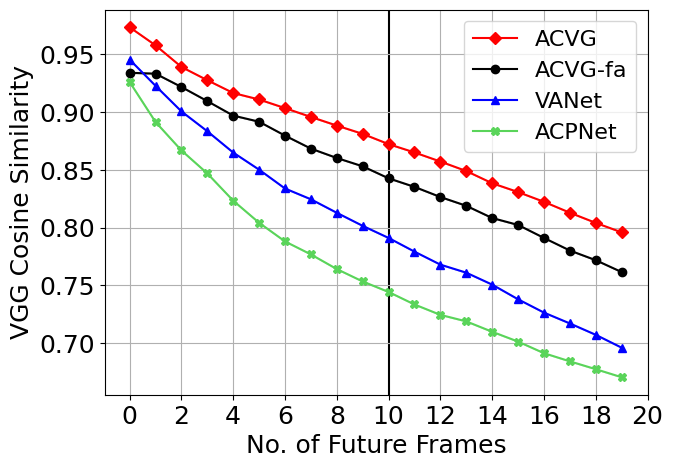}}%
\hfill 
\subcaptionbox{\label{fig:roam_quant_lpips}}{\includegraphics[width=0.25\textwidth,height=3.6cm]{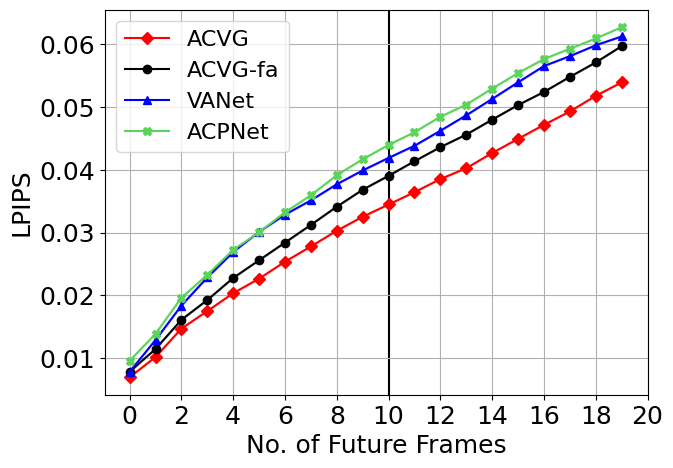}}%
\hfill 
\subcaptionbox{\label{fig:roam_quant_psnr}}{\includegraphics[width=0.225\textwidth,height=3.6cm]{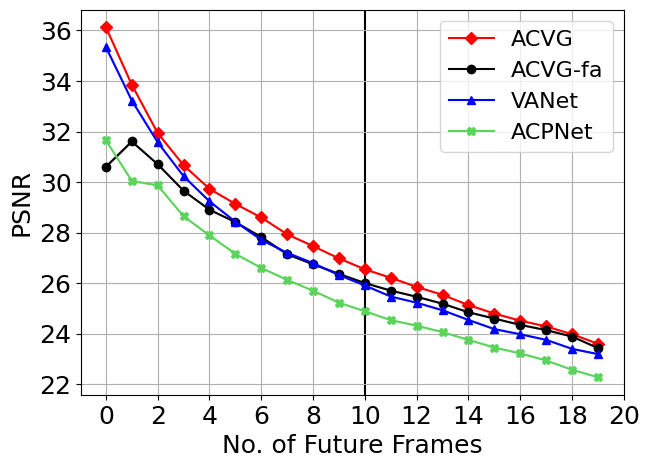}}%
\hfill 
\subcaptionbox{\label{fig:action1_error}}{\includegraphics[width=0.22\textwidth,height=3.6cm]{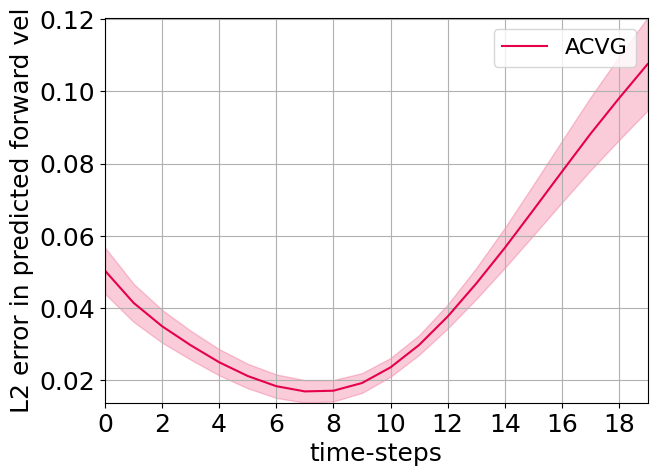}}%
\caption{A Frame-wise quantitative analysis of   ACVG, ACVG-fa, ACPNet, VANet on RoAM dataset for predicting 20 frames into the future based on the past history of 5 frames. Starting from the left, we have plotted the mean performance index for VGG 16 Cosine Similarity (\textbf{higher is better}), LPIPS score (\textbf{lower is better}), and PSNR (\textbf{higher is better}) on the test set. \cref{fig:action1_error} shows $\text{L}_2$ error between the normalised forward velocity action predicted by the Actor-network of ACVG and the ground truth. }
\label{fig:wacv_quant}
\end{figure*}
\begin{figure*}
\centering
\subcaptionbox{\label{fig:roam_acvg_half_vgg16}}{\includegraphics[width=0.25\textwidth,height=3.6cm]{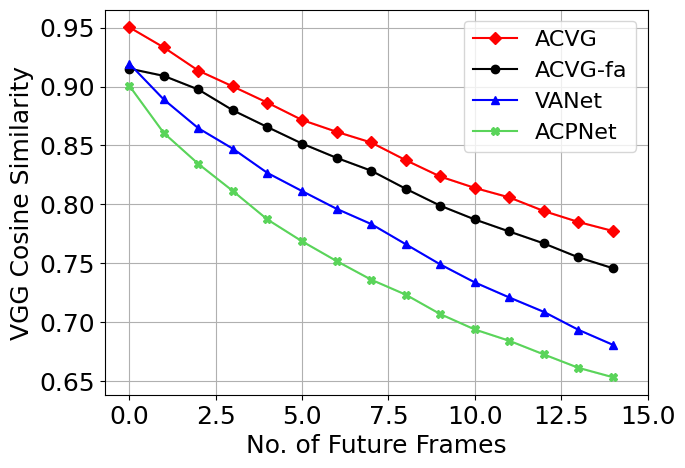}}%
\hfill 
\subcaptionbox{\label{fig:roam_acvg_half_lpips}}{\includegraphics[width=0.23\textwidth,height=3.6cm]{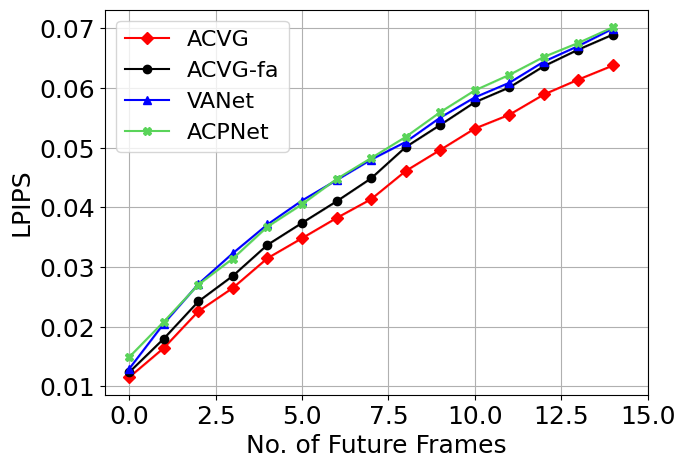}}%
\hfill 
\subcaptionbox{\label{fig:roam_acvg_dis_vgg16}}{\includegraphics[width=0.25\textwidth,height=3.6cm]{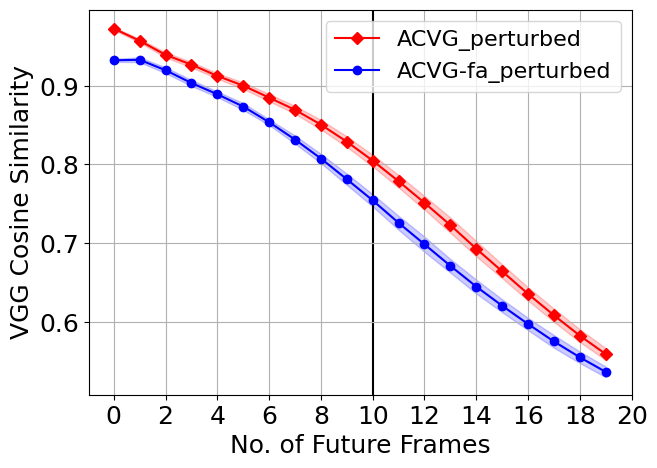}}%
\hfill 
\subcaptionbox{\label{fig:roam_acvg_dis_lpips}}{\includegraphics[width=0.23\textwidth,height=3.6cm]{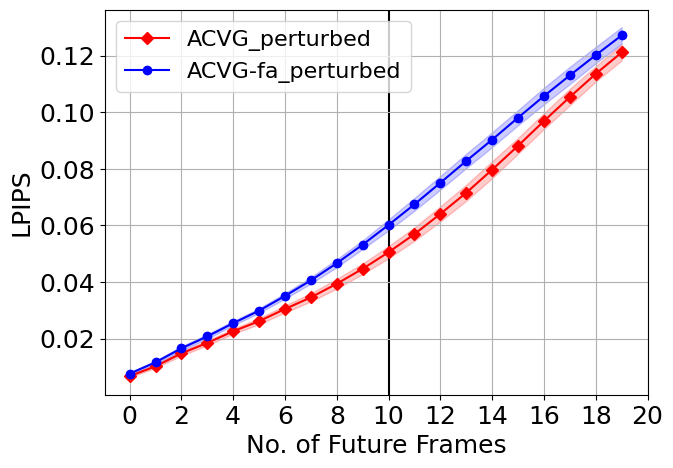}}%
\caption{A Frame-wise ablation study on  ACVG, ACVG-fa, ACPNet and VANet on RoAM dataset. Fig. \ref{fig:roam_acvg_half_vgg16} and \cref{fig:roam_acvg_half_lpips} shows the VGG 16 Cosine Similarity (higher is better) and LPIPS score (lower is better) respectively for predicting 15 frames into the future from past 5 frames at 0.5 fps$_{\text{train}}$ or $\Delta t_{\text{test}}=2\times\Delta t_{\text{train}}$ . Fig. \ref{fig:roam_acvg_dis_vgg16} and \cref{fig:roam_acvg_dis_lpips} plots the mean VGG 16 Cosine Similarity and LPIPS score with 95\% confidence for predicting 20 frames from past 5 frames in the presence of a random perturbation $\mathcal{N}(0,0.2)$ in the action value. }
\label{fig:wacv_ablation}
\end{figure*}
\begin{figure*}
\resizebox{\textwidth}{!}{\begin{tabular}{cccccccccccc}
 & t=1 & t=3& t=5  & t=7 & t=9 & t=11 & t=13 & t=15 & t=19 & t=23\\
 GT velocity& $v=0.0826$ & $v=0.0826$ & $v=0.083$ & $v=0.0833$ & $v=0.0836$ & $v=0.0836$ & $v=0.084$ & $v=0.0843$ & $v=0.0846$ & $v=0.085$\\
Ground Truth & \imgcell{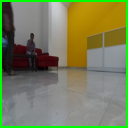} & \imgcell{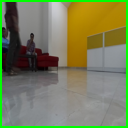} & \imgcell{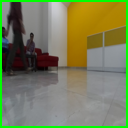}&  \imgcell{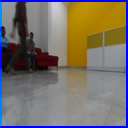}&\imgcell{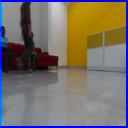} & \imgcell{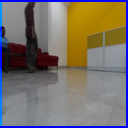} & \imgcell{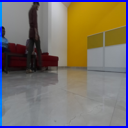} & \imgcell{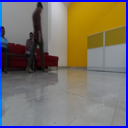} & \imgcell{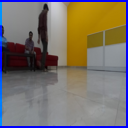} & \imgcell{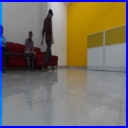} \\
&&&ACVG&  \imgcell{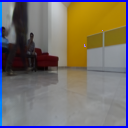}&\imgcell{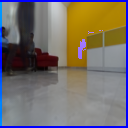} & \imgcell{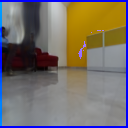} & \imgcell{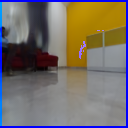} & \imgcell{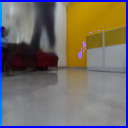} & \imgcell{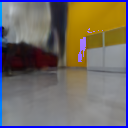} & \imgcell{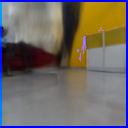} \\
&&& predicted velocity in ACVG  & $v=0.064$ & $v=0.068$ & $v=0.072$ & $v=0.075$ & $v=0.081$ & $v=0.09$ & $v=0.096$\\
&&&ACPNet&  \imgcell{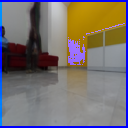}&\imgcell{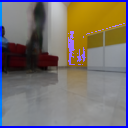} & \imgcell{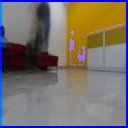} & \imgcell{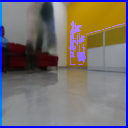} & \imgcell{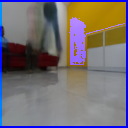} & \imgcell{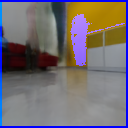} & \imgcell{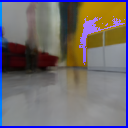} \\
&&& fixed velocity for ACVG-fa \& ACPNet  & $v=0.083$ & $v=0.083$ & $v=0.083$ & $v=0.083$ & $v=0.083$ & $v=0.083$ & $v=0.083$\\
&&&ACVG-fa&  \imgcell{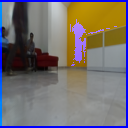}&\imgcell{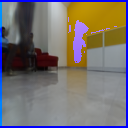} & \imgcell{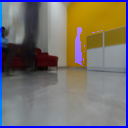} & \imgcell{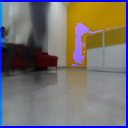} & \imgcell{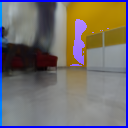} & \imgcell{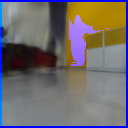} & \imgcell{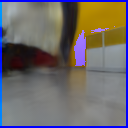}\\
&&&VANet&  \imgcell{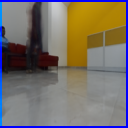}&\imgcell{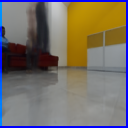} & \imgcell{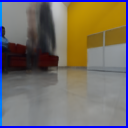} & \imgcell{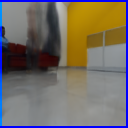} & \imgcell{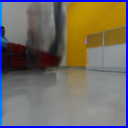} & \imgcell{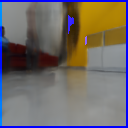} & \imgcell{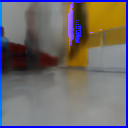}
\end{tabular}}
\caption{Predicted raw image frames along with the corresponding forward velocity values for ACVG, ACVG-fa, ACPNet and VANet on RoAM dataset for qualitative performance analysis. Models predicted 20 future frames based on the past 5 frames.}
\label{fig:wacv_qual}
\end{figure*}
 For the quantitative analysis of the performance, we have employed four evaluation metrics: Peak Signal-to-Noise Ratio (PSNR), VGG16 Cosine Similarity \cite{VGG16}, and Fréchet Video Distance (FVD) \cite{FVD} and Perceptual Metric LPIPS \cite{lpips}. Among these metrics, FVD measures the spatio-temporal distribution of the generated videos as a whole, with respect to the ground truth, based on the Fréchet Inception Distance (FID) that is commonly used for evaluating the quality of images from generative frameworks. For frame-wise evaluation, we provide comparative performance plots for VGG16 cosine similarity index, LPIPS, and PSNR. The VGG16 cosine similarity index measures the cosine similarity between flattened high-level feature vectors extracted from the VGG network \cite{VGG16}, providing insights into the perceptual-level differences between the generated and ground truth video frames. The VGG cosine similarity has emerged as a widely adopted standard for evaluating frame-wise similarity in the vision community, as seen in works like \cite{villegasNeurIPS2019}. Recently LPIPS \cite{lpips} has emerged as a popular measure \cite{slrvp2020} of human-like perceptual similarity between two image frames and uses pretrained models like Alex-net in its evaluation function. Comparatively, PSNR uses a much shallow function \cite{lpips,slrvp2020} and is known to be biased towards blurry images.\\
\textbf{Quantitative Evaluation:} The figures presenting the quantitative performance metrics for VGG16 similarity, LPIPS score, and Peak Signal-to-Noise Ratio (PSNR) are displayed in \cref{fig:roam_quant_vgg16,fig:roam_quant_lpips,fig:roam_quant_psnr} respectively where we have plotted the comparative performance among ACVG,  ACPNet, VANet along with our baseline model ACVG-fa. Unlike ACVG, ACVG with fixed action of ACVG-fa does not learn the action dynamics with an actor-network and simply considers the action of the robot to be constant during its inference phase from $t=1$ to $T$ meaning  $a_{1:T}= a_{t=0}$. The training of ACVG-fa only consists of the Generator training phase of ACVG as shown in \cref{fig:acvg_generator}. The performance plots of VGG16 cosine similarity in Fig. \ref{fig:roam_quant_vgg16}  indicate that ACVG performs much better than ACPNet,  VANet and ACVG-fa. Fig. \ref{fig:roam_quant_vgg16} illustrates that at the initial time step ($t=1$), both VANet and ACVG-fa commence with the same similarity index. However, as time progresses, VANet experiences a rise in prediction error, attributed to the compounding effect of accumulated errors from previous time steps. It is also worth noting that the parametric space for VANet is almost twice the parametric space of ACVG and ACVG-fa. In \cref{fig:roam_quant_lpips} all 4 models start at similar perception index, however, as time progresses ACVG out-performs the other 3 frameworks with considerable margins.
\begin{table}[!htb]
    \caption{FVD Score}
    \begin{subtable}{.6\linewidth}
      \centering
        \caption{$\Delta t_{\text{test}}=\Delta t_{\text{train}}$}
        \begin{tabular}{ll}
            \hline
 Model & Score  \\ \hline
 ACVG & 383.68  \\ 
 ACVG-fa & 505.015 \\ 
 VANet & \textbf{350.98} \\
 ACPNet & 490.96 \\ 
 MCNet & 625.54 \\ \hdashline
 ACVG$_{\text{perturbed}}$ & \textbf{499.77 $\pm 22$}  \\ 
 ACVG-fa$_{\text{perturbed}}$ & 661.00 $\pm 24$ \\
 \hline
\end{tabular}\label{tab:fvd1}
    \end{subtable}
    \begin{subtable}{.35\linewidth}
      \centering
        \caption{$\Delta t_{\text{test}}=2\times\Delta t_{\text{train}}$}
        \begin{tabular}{|rr}\hline
            Model & Score  \\ \hline
 ACVG & 271.4  \\ 
 ACVG-fa & 325.61 \\ 
 VANet & \textbf{247.67} \\
 ACPNet & 308.84 \\ \hline
  &  \\
 &  \\
  &  \\
 \end{tabular}\label{tab:fvd2}
    \end{subtable} 
\label{tab:fvd}
\end{table}
The comparative performance analysis among ACVG, ACVG-fa and ACPNet, as depicted in Figs.  \ref{fig:roam_quant_vgg16} -  \ref{fig:roam_quant_psnr} highlights the advantages of approximating the dual image and action pair dynamics, as outlined with Eqns. \ref{eq:predict_x_infr} and \ref{eq:actor} in \cref{sec:acvg}. Interestingly, \cref{tab:fvd1} reveals that VANet outperforms ACVG by a marginal amount in  FVD score. This is attributed to the larger parametric space of the VANet architecture compared to ACVG which leads to sharper predictions of smaller objects in the frames. However, sharper predictions do not guarantee accurate estimation of the movement of the objects in the frame which is supported by our comparative performance plots in Figs. \ref{fig:roam_quant_vgg16}, \ref{fig:roam_quant_lpips}, and \ref{fig:roam_quant_psnr}. In order to show the dual dependency between the action and generated image frames, we have also plotted the $\text{L}_2$ norm error between the predicted normalized action from the Actor-network to the ground truth values in \cref{fig:action1_error}, which shows that the Actor-network accurately predicts actions when the Generator network produces high-quality image frames. However, after $t>8$ time steps, when image quality degrades, the error in predicted actions increases.

We have performed an extensive ablation study using ACVG, ACVG with fixed action (ACVG-fa), ACPNet, and VANet to analyze the impact of altering the sampling time-step and introducing random perturbations to the action input of the Generator module. The quantitative results are depicted in \cref{fig:wacv_ablation}.  In order to study the effect of changing the sampling interval on the performances of the models, during inference we use a test sampling interval that is twice the train sampling interval, or, $\Delta t_{\text{test}}=2\times\Delta t_{\text{train}}$. The VGG16 cosine similarity and LPIPS for the prediction of 15 future frames from 5 past frames are given in Figs. \ref{fig:roam_acvg_half_vgg16} and \ref{fig:roam_acvg_half_lpips} and we observe that ACVG outperforms all 3 frameworks with a significant margin. More interestingly, in \cref{fig:roam_acvg_half_lpips} we observe that both ACVG and ACVG-fa start from the same point but as time progresses, ACVG-fa's performance degrades and it behaves more like ACPNet. \cref{tab:fvd2} shows the FVD scores from all the 4 frameworks inferring at fps$_{\text{test}}=$ 0.5 fps$_{\text{train}}$. It should be noted that in \cref{tab:fvd2} scores are evaluated for only 15 predicted frames compared to the \cref{tab:fvd1} where the no of the predicted image frame is 20 which is why the scores are relatively lower.

To study the effect of perturbation to action input on the generated image frames we added a random noise, $\mathcal{N}(0,0.2)$ to the normalised action inputs of ACVG and ACVG-fa and plotted their mean VGG16 cosine similarity and LPIPS scores in \cref{fig:roam_acvg_dis_vgg16,fig:roam_acvg_dis_lpips} respectively. \cref{fig:roam_acvg_dis_vgg16,fig:roam_acvg_dis_lpips} shows that the addition of random noise to the action input adversely affects the performance of both the models, however, ACVG  continues to outperform ACVG-fa with a clear margin. The same can be observed from their FVD scores in the \cref{tab:fvd1} too. This further validates our hypothesis that action improves the prediction accuracies of generated image frames in partially observable scenarios. 

\noindent\textbf{Qualitative Evaluation:} Fig. \ref{fig:wacv_qual} contains examples of raw image frames along with the input forward velocity of the camera from the test set, generated by ACVG, ACPNet, VANet, and ACVG-fa along with the first row representing the ground truth values. 
 Closely observing the gait of the human motion in Fig. \ref{fig:wacv_qual} from $t=9 \text{ to } 23$, we can clearly see how ACVG does a much better job at approximating the motion of the human figure compared to the other three frameworks. 
 From time-steps $t=9, \text{ to } 23$ for ACPNet and VANet produces considerably blurry predictions. ACVG-fa initially does a better job compared to ACPNet and VANet however from $t=13$, it also produces blurred outputs of the human gait. These findings are in coherence with our quantitative results reported in \cref{fig:wacv_quant}.
 
\cref{fig:wacv_quant,fig:wacv_ablation} provide empirical validation to our hypothesis that in scenarios with partial observability, accurate modeling of camera action dynamics enhances accuracy in predicting image frames and vice versa. Furthermore, while we validated our framework on RoAM dataset which is recorded on a Trutlebot3 robot, it should be noted that during training we only used the normalised control actions of the robot thus these results can be extended and generalised to other wheeled robots. 

\section{Conclusion}\label{sec:conclusion}
A new video prediction framework (ACVG), capable of generating multi-timestep future image frames in partially observable scenarios, is presented. It explored learning the interplay of dynamics between the observed image and robot action pair with its dual Generator Actor architecture. The detailed empirical results demonstrate that modeling the dynamics of control action data based on image state information yields superior approximations of predicted future frames. Notably, ACVG establishes a causal link between two distinct data modalities: the low-dimensional control action data $a_t$ from the robot and the high-dimensional observed image state $x_t$. This has potential applications in reinforcement learning frameworks.

{
    \small
    \bibliographystyle{ieeenat_fullname}
    \bibliography{egbib}
}


\end{document}